\documentclass[nohyperref]{article}

\usepackage{microtype}
\usepackage{graphicx}
\usepackage{booktabs} % for professional tables
\usepackage{floatrow}

\usepackage{hyperref}

\usepackage[accepted]{icml2022}

\usepackage{amsmath}
\usepackage{amssymb}
\usepackage{mathtools}
\usepackage{sidecap}
\usepackage{subcaption}
\usepackage{wrapfig}
\usepackage{amsthm}
\usepackage{microtype}
\usepackage{graphicx}
\usepackage{booktabs} % for professional tables
\usepackage{footnote}
\usepackage{hyperref}

\usepackage{amsmath}
\usepackage{amssymb}
\usepackage{enumitem}
\usepackage{graphicx}
\usepackage{xspace}
\usepackage{array,multirow}
\usepackage{xargs} % commandx
\usepackage{url}
\usepackage{ragged2e}
\usepackage{bbm}
\usepackage{bm}
\usepackage{multirow}
\usepackage[normalem]{ulem}
\usepackage{xcolor}
\usepackage{wrapfig}
\definecolor{mygreen}{HTML}{8DD3C7}
\usepackage{tikz}
\newcommand{\tikzcircle}[2][red,fill=red]{\tikz[baseline=-0.75ex]\draw[#1,radius=#2] (0,0) circle ;}%
\usetikzlibrary{shapes.geometric, calc}

\newcommand{\roberta}{\textsc{Roberta}\xspace}

\newcommand{\webqsp}{WebQSP\xspace}
\newcommand{\fbqa}{FreebaseQA\xspace}
\newcommand{\metaqa}{MetaQA\xspace}

\newcommand{\alg}{\textsc{Cbr-subg}\xspace}
\newcommand{\knnq}{$\mathrm{kNN}_q$\xspace}

\newcommand{\ignore}[1]{}

\newlength{\mysize}

\usepackage[capitalize,noabbrev]{cleveref}

\theoremstyle{plain}

\theoremstyle{definition}

\theoremstyle{remark}

\usepackage[textsize=tiny]{todonotes}
\DeclareMathOperator*{\argmax}{arg\,max}

\newcommand\score[2]{%
  \pgfmathsetmacro\pgfxa{#1 + 1}%
  \tikzstyle{scorestars}=[star, star points=5, star point ratio=2.25, draw, inner sep=0.15em, anchor=outer point 3]%
  \begin{tikzpicture}[baseline]
    \foreach \i in {1, ..., #2} {
      \pgfmathparse{\i<=#1 ? "yellow" : "gray"}
      \edef\starcolor{\pgfmathresult}
      \draw (\i*1em, 0) node[name=star\i, scorestars, fill=\starcolor]  {};
    }
    \pgfmathparse{#1>int(#1) ? int(#1+1) : 0}
    \let\partstar=\pgfmathresult
    \ifnum\partstar>0
      \pgfmathsetmacro\starpart{#1-(int(#1)}
      \path [clip] ($(star\partstar.outer point 3)!(star\partstar.outer point 2)!(star\partstar.outer point 4)$) rectangle 
      ($(star\partstar.outer point 2 |- star\partstar.outer point 1)!\starpart!(star\partstar.outer point 1 -| star\partstar.outer point 5)$);
      \fill (\partstar*1em, 0) node[scorestars, fill=yellow]  {};
    \fi
  \end{tikzpicture}%
  }
  
\icmltitlerunning{Case-based Reasoning for KBQA over Subgraphs}

\begin{document}

\twocolumn[
% \icmltitle{Semiparametric Subgraph Reasoning \\ for Question Answering over Large Knowledge Bases}
\icmltitle{Knowledge Base Question Answering by \\ Case-based Reasoning over Subgraphs}

\icmlsetsymbol{equal}{*}

\begin{icmlauthorlist}
\icmlauthor{Rajarshi Das}{equal,umass}
\icmlauthor{Ameya Godbole}{equal,usc}
\icmlauthor{Ankita Naik}{umass}
\icmlauthor{Elliot Tower}{umass}
\icmlauthor{Robin Jia}{usc}\\
\icmlauthor{Manzil Zaheer}{deepmind}
\icmlauthor{Hannaneh Hajishirzi}{uw}
\icmlauthor{Andrew McCallum}{umass}
\end{icmlauthorlist}

\icmlaffiliation{umass}{UMass Amherst}
\icmlaffiliation{usc}{University of Southern California}
\icmlaffiliation{uw}{University of Washington}
\icmlaffiliation{deepmind}{Google DeepMind}

\icmlcorrespondingauthor{Rajarshi Das}{rajarshi@cs.washington.edu}
\icmlcorrespondingauthor{Ameya Godbole}{ameyagod@usc.edu}

\icmlkeywords{Machine Learning, ICML}

\vskip 0.3in
]

\printAffiliationsAndNotice{\icmlEqualContribution} % otherwise use the standard text.

\begin{abstract}

Question answering (QA) over knowledge bases (KBs) is challenging because of the diverse, essentially unbounded, types of reasoning patterns needed.
However, we hypothesize in a large KB, reasoning patterns required to answer a query type reoccur for various entities in their respective subgraph neighborhoods.
Leveraging this structural similarity between local neighborhoods of different subgraphs, we introduce a semiparametric model (\alg) with 
(i) a nonparametric component that for each query, dynamically retrieves other similar $k$-nearest neighbor (KNN) training queries along with query-specific subgraphs and 
(ii) a parametric component that is trained to identify the (latent) reasoning patterns from the subgraphs of KNN queries and then apply them to the subgraph of the target query. 
We also propose an adaptive subgraph collection strategy to select a query-specific compact subgraph, allowing us to scale to full Freebase KB containing billions of facts. We show that \alg can answer queries requiring subgraph reasoning patterns and performs competitively with the best models on several KBQA benchmarks. Our subgraph collection strategy also produces more compact subgraphs (e.g.  55\% reduction in size for WebQSP while increasing answer recall by 4.85\%)\footnote{Code, model, and subgraphs are available at \url{https://github.com/rajarshd/CBR-SUBG}}.
\vspace{-3mm}

\end{abstract}

\section{Introduction}
\label{sec:intro}
\begin{figure}
    \centering
    \includegraphics[width=\columnwidth]{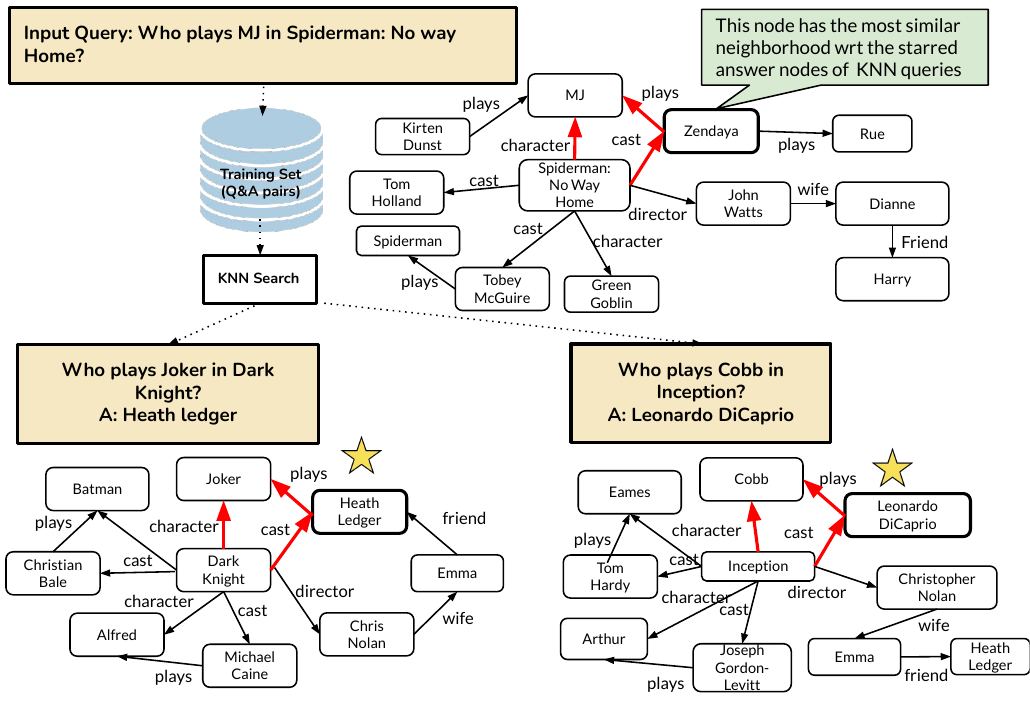}
    \caption{Figure shows an input query and two queries in the training set that are similar to the input query. 
    The relevant subgraph for each query (query subgraph) is also shown. Note that the ``reasoning patterns'' required to answer the queries (edges in \textcolor{red}{red}) repeats in the subgraphs of each query. 
    Also note, the corresponding answer nodes (marked as \score{1}{1}) are analogously located within the reasoning patterns in each subgraph. 
    Thus the answer node can be found by identifying the node in the query subgraph that is most similar to the answer nodes in the subgraph of KNN queries.}
    \label{fig:intro}
    % \vspace{-5mm}
\end{figure}

Knowledge bases (KBs) store massive amounts of rich symbolic facts about real-world entities in the form of relation triples---$\mathrm{(e_1, r, e_2})$, where $e_1, e_2$ denote entities and $r$ denotes a semantic relation. 
KBs can be naturally described as a graph where the entities are nodes and the relations are labelled edges. 
An effective and user-friendly way of accessing the information stored in a KB is by issuing queries to it. 
Such queries can be structured (e.g. queries for booking flights) or unstructured (e.g. natural language queries). 
The set of KB facts useful for  answering a query induce a reasoning pattern---e.g. a chain of KB facts forming a path or more generally a subgraph in the knowledge graph (KG) (set of \textcolor{red}{red} edges in Figure~\ref{fig:intro}). 
It is very laborious to annotate the reasoning patterns for each query at scale and hence it is important to develop weakly-supervised knowledge base question answering (KBQA) models that do not depend on the availability of the annotated reasoning patterns.

We are interested in developing models that can answer queries requiring complex subgraph reasoning patterns.
Many previous works in KBQA \cite{neelakantan2015compositional,deeppath,das2018go} reason over individual relational paths. 
However, many queries require a model to reason over a set of facts \emph{jointly}. 
For example, the query in Figure~\ref{fig:intro} cannot be answered by considering individual paths. 
Furthermore, a model has to learn a large number of reasoning patterns because of the diverse nature of possible questions. 
Moreover, a model may encounter very few examples of a particular pattern during training, making it challenging for the models to learn and encode the patterns entirely in its parameters.
A possible solution to this challenge might lie in a classical AI paradigm proposed decades back---Case-based Reasoning (CBR) \citep{schank1982dynamic}. 
In a CBR framework, a new problem is solved by retrieving other \emph{similar} problems and reusing their solution to derive a solution for the given problem. 
In other words, models, instead of memorizing patterns in its parameters, can instead reuse the reasoning patterns of other similar queries, retrieved dynamically during inference. 
Recently, CBR was successfully used for KB completion by \citet{das2020simple,das2020probabilistic}.

This paper introduces a semiparametric CBR-based model (\alg) for QA over KBs  with a nonparametric component, that for each query, dynamically retrieves other similar $k$-nearest neighbor (KNN) queries from the training set. 
To retrieve similar queries, we use masked sentence representation of the query \cite{soares2019matching} obtained from pre-trained language models. 

We hypothesize that the reasoning patterns required for answering similar queries reoccur within the subgraph neighborhood of entities present in those queries (Figure~\ref{fig:intro}). 
The answer nodes for each query are also analogously nestled within the reasoning patterns (marked as \score{1}{1} in Figure~\ref{fig:intro}) of the query subgraphs, i.e. they have similar neighborhoods. 
However, we do not have annotated reasoning patterns that could be used to search for the answer node. 
Moreover, a subgraph can have tens of thousands of entity nodes. How do we still identify the answer nodes in the query subgraph?

To identify the answer nodes, our model has a parametric component comprising of a graph neural network (GNN) that is trained to identify the (latent) reasoning patterns from the subgraphs of KNN queries and apply it to the subgraph of the target query. 
GNNs have been shown to be effective in encoding structural properties of local neighborhoods in the node representations~\citep{duvenaud2015convolutional,kipf2017semi}. 
We leverage node representations obtained from GNNs for finding answer nodes. 
Specifically, the answer nodes are identified by performing a nearest neighbor search for the most similar nodes in the query subgraph w.r.t the representation of answer nodes in the KNN subgraph. 
The parametric model is trained via contrastive learning (\S\ref{sub:gnn}) \cite{chopra2005learning,gutmann2010noise}. 

A practical challenge for KBQA models is to select a compact subgraph for a query. 
The goal is to ensure that the subgraph has high recall and is small enough to fit into GPU memory for gradient-based learning. 
Many KBQA methods usually consider few hops of edges around entities as the query subgraph \cite{neelakantan2015compositional,saxena2020improving} leading to query-independent and (often) large subgraphs, because of the presence of hub nodes in large KBs. 
We propose an \emph{adaptive} subgraph collection method tailored for each query where we use our nonparametric approach of retrieving KNN queries to help gather the query subgraph leading to compact subgraphs with higher recall of reasoning patterns (\S~\ref{sub:subgraph_retrieval}).

An important property of nonparametric models is its ablility to grow and reason with new data. 
Being true to its nonparametric design, \alg uses sparse representations of entities that makes it easy to represent new entities. 
Moreover, we also demonstrate that the performance of \alg improves as more evidence is retrieved, suggesting that \alg can reason with new evidence.

\textbf{Contributions.} To summarize, this paper introduces \alg, a semiparametric model for weakly-supervised KBQA that retrieves similar queries and utilizes the similarities in graph structure of local subgraphs to answer a query (\S\ref{sub:gnn}). 
We also propose a practical algorithm for gathering query-specific subgraphs that utilizes the retrieved KNN queries to produce compact query-specific subgraphs (\S\ref{sub:subgraph_retrieval}). 
We show that \alg can model (latent) subgraph reasoning patterns (\S\ref{sub:subgraph_pattern}), more effectively than parametric models; can reason with new entities (\S\ref{sub:subgraph_pattern}) and new evidence (\S\ref{sub:knn_increase}). 
Lastly, we perform competitively with state-of-the-art KBQA models on multiple benchmarks. For example, on the FreebaseQA dataset \cite{jiang2019freebaseqa}, we outperform most competitive baseline by 14.45 points.

\section{Related Work}
\label{sec:related_work}
\textbf{CBR-based models for KB reasoning.} Recently, \citet{das2020simple,das2020probabilistic} proposed a CBR-based technique for KB completion.
However, their work has several limitations. Firstly, it can only model simple linear chains.
Secondly, it uses exact symbolic matching to find similarities in patterns between cases and the query.
Lastly, it cannot handle natural language queries and only works with structured slot-filling queries.
In contrast, \alg can model arbitrary reasoning patterns; uses soft-matching by comparing representations of answer nodes and can handle natural language queries. Lastly, our method outperforms \citet{das2020simple} on various benchmarks.
A follow up work of \citet{das2021case} proposed a CBR model that can handle natural language queries, however that work requires the availability of annotated reasoning patterns for training, an important distinction from our work that does not need any annotation of reasoning patterns.

\noindent\textbf{Semiparametric models for KBQA.} GraftNet \cite{sun2018open} and PullNet \cite{sun2019pullnet} are two semiparametric models for KBQA where they like us, provide both a mechanism of collecting a query-subgraph and reasoning over them.
For their nonparametric component, these work employ a retrieval process where a parametric model classifies which edges would be relevant to the query.
Being parametric, these models cannot generalize to new type of questions without re-training the model parameters.
However, our nonparametric approach will work as it retrieves similar queries on-the-fly.
For their reasoning model, both works use a graph convolution model and treat the answer prediction as a node classification task.
However, unlike us they do not reason with subgraphs of similar KNN queries.
Lastly, we compare extensively with them and outperform them on multiple benchmarks.
Our approach also has similarities with retriever-reader architecture for open-domain QA over text \cite{chen2017reading} where a retriever selects evidence specific to the query and the reader reasons with them to produce the answer.
 
\noindent\textbf{Graph representations using contrastive learning.} Recently, there has been a lot of work on learning graph representations via contrastive learning \cite{hassani2020contrastive,you2020graph,sun2019infograph,qiu2020gcc,zhu2020deep} where they create two views of the same graph by randomly dropping edges and nodes.
Next, the two views of subgraphs are treated as positive pair and their representations are contrasted wrt other negative samples.
Our work differs from them because we do not create different views of the same graph, rather we follow the CBR hypothesis and make node representations of answer nodes of two query-specific subgraphs similar provided the queries have relational similarity.
 
 \noindent\textbf{Semantic parsing.} Classic works in semantic parsing \cite{zelle1996learning,zettlemoyer2012learning,zettlemoyer2007online} were among the early works to use statistical learning to convert queries to executable logical forms.
However, these work needed annotated logical forms during training.
Follow up work \cite{berant2013semantic,kwiatkowski2013scaling} learned semantic parser directly from question answer pairs.
However, their model relied on simple hand crafted features.
Recent approaches to semantic parsing \cite{saxena2020improving,he2021improving} uses powerful neural models and achieve strong performance.
However unlike us, these parametric models learn dense representation of entities and hence will not generalize to unseen entities like our approach.
 
 \noindent\textbf{Inductive KB reasoning.} Our model is also related to \citet{teru2020inductive} as it explores KB reasoning in an inductive setting.
They also have a sparse representation of entities.
However, the task they consider is predicting a KB relation between two nodes which is an easier task than performing KBQA using natural language queries.
 
 \noindent\textbf{Graph neural networks.}
 A model like \alg is possible for KBQA because of tremendous progress made in graph representation learning by message passing neural networks \citep[inter-alia]{kipf2017semi,Velickovic2018GraphAN,schlichtkrull2018modeling}.
\alg is not dependent on any specific message passing scheme and can work with any GNN architecture that can operate over heterogenous knowledge graphs.
For our experiments, we use the widely used relational-GCN \cite{schlichtkrull2018modeling}.
Further related work is included in the appendix (\S\ref{sec:further_related_work}).

\vspace{-2mm}
\section{Model}
\vspace{-1mm}
\label{sec:model}
\begin{figure*}
    \centering
    \includegraphics[width=\columnwidth]{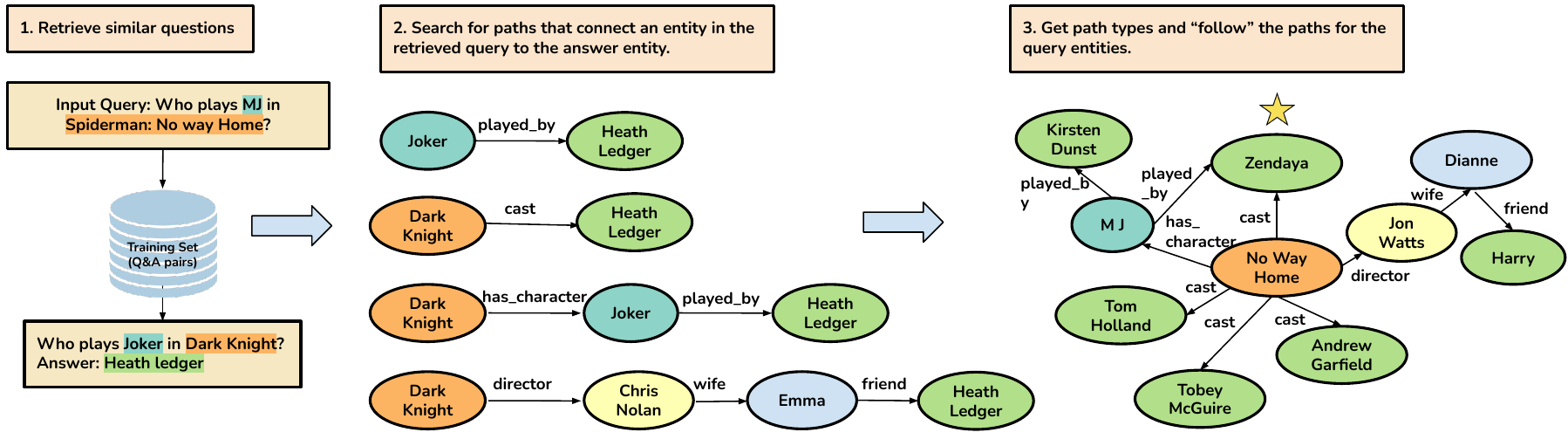}
    \vspace{-1mm}
    \caption{Figure shows the query-subgraph selection procedure with 1-nearest neighbor retrieved query. Graph paths connecting the entities in the retrieved query and its answer are collected. Next the sequence of relations (path types) are gathered and are then followed starting from the entity in the given query. All the edges spanned by this process are collected to form the query-specific subgraph. This process is repeated for each of the $k$ retrieved queries.}
    \label{fig:subgraph_collection}
    % \vspace{-10mm}
\end{figure*}

This section describes the nonparametric and parametric components of \alg. 
In CBR, a case is defined as an abstract representation of a problem along with its solution. 
In our KBQA setting, a case is a natural language query along with its answer. 
Note in KBQA, answers are entities in the KB (or nodes in KG).

\textbf{Task Description}. Let $q$ be a natural language query and let $\mathcal{K}$ be a symbolic KG that needs to be queried to retrieve an answer list $\mathcal{A}$ containing the answer(s) for $q$. 
We assume access to a training set $\mathcal{D}$ = $\{(q_1, \mathcal{A}_1), (q_2, \mathcal{A}_2), \ldots (q_N, \mathcal{A}_N)\}$ of query-answer pairs where $\mathcal{A}_i$ denote the list of answer nodes for $q_i$. 
The training set $\mathcal{D}$ forms the `case-base'. 
The reasoning pattern (a.k.a. logical form) required to answer a query are the set of KG edges required to deduce the answer to $q$. Let $P_q$ denote this set of edges. 
For example, in Figure~\ref{fig:intro}, for $q =$ ``Who plays `MJ' in No Way Home?'', $P_q$ = \{(\textrm{MJ, played\_by, Zendaya}), (\textrm{No Way Home, has\_character, MJ}), \textrm{(No Way Home, cast, Zendaya)}\}. 
We define reasoning pattern `type' for a pattern as the set of edges where the entities have been replaced by free variables. For example, $T(P_{q})$ = \{(\textrm{M, played\_by, Z}), (\textrm{S, has\_character, M}), \textrm{(S, cast, Z)}\}. 
It should be noted that \alg does not assume access to annotated $P_q$.

\textbf{Method overview}. For input $q$ and $\mathcal{K}$, \alg first retrieves $k$ similar query-answer(s) w.r.t $q$ from $\mathcal{D}$ (\S~\ref{sub:knn_query_retrieval}). 
Denote this retrieved set as $\mathrm{kNN}_q \subset \mathcal{D}$. Next, \alg finds query-specific subgraphs $\mathcal{K}_{q_{i}}$ for each query in $\{q\} \cup $ \knnq (\S~\ref{sub:subgraph_retrieval}). 
According to the CBR hypothesis, the reasoning required to solve a new problem will be similar to the reasoning required to solve other similar problems. 
Similarly for our KBQA setting, we hypothesize that the reasoning pattern type, $T(P_{q})$ repeats across the neighborhood of query subgraphs of \knnq. 
Next, \alg uses graph neural networks (GNNs) to encode local structure into node representations (\S~\ref{sub:gnn}). 
Now, if the CBR hypothesis holds true, then the representation of the answer nodes in each query subgraphs will be similar as the local structure around them share similarities. 
Hence, the answer node of the given query $q$ can be identified by searching for the node that has the most similar representations to the answer nodes in the query subgraphs of $\mathrm{kNN}_q$.

\subsection{Retrieval of Similar Cases}
\label{sub:knn_query_retrieval}
Given the input query $q$, \alg first retrieves other similar cases from the training set. 
We represent the query by embeddings obtained from large pre-trained language models \cite{liu2019roberta}. 
Inspired by the recent advances in neural dense passage retrieval \cite{karpukhin2020dense}, we use a \roberta-base encoder to encode each question independently. 
A single representation is obtained by mean pooling over the token-level representations.

Generally, we want to retrieve questions that express similar relations rather than retrieving questions that are about similar entities. 
For example, for the query, `Who played Natalie Portman in Star Wars?', we would like to retrieve queries such as `Who played Ken Barlow in Coronation St.?' instead of `What sports does Natalie Portman like to play?'. 
To obtain entity-agnostic representation, we replace the entity spans with a special `\textsc{[mask]}' token in the query, i.e. the original query becomes `Who played \textsc{[mask]} in \textsc{[mask]}?'. 
The entity masking strategy has previously been successfully used in learning entity-independent relational representations \cite{soares2019matching}. 
The similarity score between two queries is given by the inner product between their normalized vector representations (cosine similarity). 
We pre-compute the representations of queries in the train set. 
During inference, the most similar query representations are obtained by doing a nearest neighbor search over the representations stored in the case-base.

\subsection{Query-subgraph Selection}
\label{sub:subgraph_retrieval}

A practical challenge for KBQA models is to select a subgraph around the entities present in the query. 
The goal is to ensure that the necessary reasoning patterns and answers are included while producing a graph small enough to fit into GPU memory for gradient-based learning. 
A na\"ive strategy to select a subgraph is to consider all edges in 2-3 hops around the query entities. 
This strategy leads to subgraphs which are independent of the query.
Moreover, in large KGs like Freebase \cite{fb}, considering the full 2 or 3-hop subgraph often leads to accumulation of millions of edges because of the presence of hub nodes.

We propose a nonparametric approach of query subgraph collection that utilizes the retrieved cases, \knnq, from the last step (Figure~\ref{fig:subgraph_collection}). 
For each of the retrieved case, chains of edges (or paths) in the graph that connect the entity in the retrieved query to its answers are collected by doing a depth-first search. 
(Note, since the retrieved queries are from the training set, we know the answer to them). 
Next, the sequence of relations are collected from each chain and they are followed starting from the entities of the input query. 
If a chain of relations do not exist from the query, then they are simply ignored. 
This process is repeated for each of the retrieved cases.
Note that, not all chains collected from the nearest neighbors are meaningful for the query. 
For example, the last (3-hop) chain collected in Figure~\ref{fig:subgraph_collection} is not relevant for answering the query and even though it ended at the answer for the retrieved query, the same is not the case for the input query. 
Such paths are often referred to as spurious paths or evidence \cite{he2021improving}. 
All the edges gathered by following this process form the subgraph for the input query. 
The underlying idea behind the subgraph selection procedure is simple --- since the paths connect queries and answers of similar queries, they should also be relevant for answering the given query.

There is a class of prior work such as Graft-Net \cite{sun2018open} and PullNet \cite{sun2019pullnet} that learn a parametric model to choose a query-specific subgraph. 
These models employ a retrieval process where a parametric model classifies which edges would be relevant to the query. 
Being parametric, these models cannot generalize to new type of questions without re-training the model parameters. 
However, our nonparametric approach will work as long as it has access to similar queries, which it can retrieve on-the-fly. 
Our subgraph selection procedure is similar to the approach proposed in \cite{das2020simple}. 
However, \citet{das2020simple} do not use this approach to collect a query-specific subgraph. 
Rather it uses each of the paths to \emph{independently} predict the answer to a query. 
In contrast, we collect all edges in the path to form a subgraph and then reason jointly over the subgraph of the query as well as the subgraph of retrieved cases as detailed in the next sub-section.

\subsection{Reasoning over Multiple Subgraphs}
\label{sub:gnn}
This section describes how \alg reasons across the subgraphs of the given query and the subgraphs of the retrieved cases. 
We use graph neural networks (GNNs) \cite{scarselli2008graph} to encode the local structure into the node representations of each subgraph. 
During training, the answer node representations of different subgraphs are made more similar to each other in comparison to other non-answer nodes. 
Inference reduces to searching for the most similar node in the query subgraph w.r.t the answer nodes in the KNN-subgraphs. 

Modern GNNs employ a neighborhood aggregation strategy (message passing) where a node representation is iteratively updated by aggregating representations from its neighbors \cite{gilmer2017neural}. 
Let $\mathcal{G}_q = (V_q, E_q)$ represent the subgraph for a query $q$ obtained from (\S\ref{sub:subgraph_retrieval}). 
Let $\bm {X}_{v}$ denote the node feature vectors for each $v \in V_q$. 

\textbf{Input node representations.} 
A property of nonparametric models is its ability to represent, reason and grow with new data. 
Knowledge graphs store facts about the world and as the world evolves, new entities and facts are added to the KG. 
Models developed for KG reasoning \citep[inter-alia]{bordes2013translating,schlichtkrull2018modeling,sun2019rotate} learn dense representations of a fixed vocabulary of entities and are hence unable to handle new entities added to the KG. 
Following our nonparametric design principles, each entity node is represented as a sparse vector of its outogoing edge (relation) types, i.e. $\bm {x}_{v} \in \{0, 1\}^{|\mathcal{R}|}$ where $\mathcal{R}$ denotes the set of relation types in the KG.
If entity $x_v$ has m distinct outgoing edge types, then the dimension corresponding to those types are set to 1. 
This is an extremely simple and flexible way of representing entities which also expresses the local structural information around each node. 
Also note that, as new entities are added or new facts are updated about an entity, the sparse representation makes it very easy to represent new entities or update existing embeddings.

\textbf{Relative distance embedding.} Each query-specific subgraph $\mathcal{G}_q$ has a few special entities --- the entities present in the input query. 
This is because the reasoning pattern is usually in the immediate subgraph surrounding the query entity. 
We treat the query entities as `center' entities and append a relative distance embedding to every other node in $\mathcal{G}_q$ \cite{zhang2018link,teru2020inductive}. 
Specifically, for each node, the representation $\bm {x}_v$ is appended with an one-hot distance embedding $\bm {x_d} \in \{0,1\}^{|d|}$ where the component corresponding to the shortest distance from the query entity is set to 1. 
In practice, we consider subgraphs upto 3-hops from the query entities, i.e. $d=4$. For queries with multiple query entities, the minimum distance is considered.

\textbf{Message passing.} Our GNN uses the graph structure and the sparse input node features $\bm {X}_{v}$ to learn intermediate node features capturing the local structure within them. 
We follow the general message-passing scheme where a node representation is iteratively updated by combining it with aggregation of it’s neighbors’ representation \cite{xu2018powerful}. 
In particular, the $l^{th}$ layer of a GNN is,
\begin{align}
    %\label{eq:neighbor_agg}
    \bm {a}_v^{l} &= \textsc{AGGREGATE}^{l} \left( \left\lbrace \bm {h}_s^{l-1}  : s \in \mathcal{N}(v) \right\rbrace, \bm {h}^{l-1}_v\right), \\
    %\label{eq:combine}
    \bm {h}_v^{l}   &= \textsc{COMBINE}^{l} \left( \bm {h}_v^{l-1}, \bm {a}_v^{l} \right),
\end{align}
where, $\bm {a}_v^{l}$ denote the aggregated message from the neighbors of node $v$, $\bm {h}_v^{l}$ denote the node representation of node $v$ in the $l$-th layer and $\mathcal{N}(v)$ denotes the neighboring nodes of $v$. 
Since KGs are heterogenous graphs with labelled edges, we adopt the widely used multi-relational R-GCN model model \cite{schlichtkrull2018modeling} which defines the aggregate step as: $\bm {a}_v^{l} =
    \sum_{r=1}^{\mathcal{R}}\sum_{s \in \mathcal{N}_r(v)}\frac{1}{|\mathcal{N}_r(v)|}W_{r}^{l}h_s^{l-1}$ and the combine step as $h_v^{l} = \text{ReLU}(W_{\textrm{self}}^{l}h_v^{l-1} + \bm {a}_v^{l})$.
For each answer node, we consider the representation obtained from the last layer.

\textbf{Training.} Let $a_{i}, a_{j}$ be an answer node in the corresponding query-subgraphs of $q_i$ and $q_j$ (i.e. $\mathcal{G}_{q_i}$, $\mathcal{G}_{q_j}$) respectively. 
Let $\mathrm{sim}(\bm a_{i},\bm a_{j}) = \bm a_{i}^\top \bm a_{j} / \lVert\bm a_{i}\rVert \lVert\bm a_{j}\rVert$ denote the inner product between $\ell_2$ normalized answer representations (i.e. cosine similarity). 
In general there can be multiple answer nodes for a query. Let $\mathcal{A}_j$ denote the set of all answer nodes for query $q_j$ in its subgraph $\mathcal{G}_{q_j}$. 
Let $\mathrm{sim}(\bm a_i, \mathcal{A}_j) = \frac{1}{\lvert\mathcal{A}_j\rvert} \sum_{a_j \in \mathcal{A}_j} \mathrm{sim}(\bm a_{i}, \bm a_{j})$, i.e. $\mathrm{sim}(\bm a_i, \mathcal{A}_j)$ represents the mean of the scores between $a_{q_i}$ and all answer nodes in $\mathcal{G}_{q_j}$. 
We aggregate the similarity score from all retrieved queries $\mathrm{kNN}_{q_i}$ for the current query $q_i$.

The loss function we use is,
\begin{equation*}
% \label{eq:loss}
    -\log \frac{\sum_{a_i \in \mathcal{A}_i} \exp( \sum_{q_j \in \mathrm{KNN}_{q_i}} \mathrm{sim}(\bm a_{i}, \mathcal{A}_j)/\tau )}{\sum_{x_{i} \in \mathcal{V}(\mathcal{G}_{q_i})} \exp(\sum_{q_j \in \mathrm{KNN}_{q_i}} \mathrm{sim}(\bm x_i, \mathcal{A}_j)/\tau)}~,
\end{equation*}
 where, $\mathcal{A}_j$ denotes the set of all answer nodes in $\mathcal{G}_{q_j}$ for a $q_j \in \mathrm{kNN}_{q_i}$, $x_i$ goes over all nodes in query-subgraph $\mathcal{G}_{q_i}$ and $\tau$ denotes a temperature parameter. 
 In other words, the loss encourages the answer nodes in $\mathcal{G}_{q_i}$ to be scored higher than all other nodes in $\mathcal{G}_{q_i}$ w.r.t the answer nodes in the retrieved query subgraphs. 
 This loss is an extension of the the normalized temperature-scaled cross entropy loss (\textit{NT-Xent}) used in \citet{chen2020simple}.\\
 \textbf{Inference.} During inference, message passing is run over each of the query-subgraph $\mathcal{G}_{q_i}$ and the subgraphs $\mathcal{G}_{q_j}$  of its $k$ retrieved queries $q_j \in \mathrm{kNN}_{q_i}$ to obtain the node representations and the highest scoring node in $\mathcal{G}_{q_i}$ w.r.t all the answer nodes in the retrieved query sub-graphs is returned as the answer.
\begin{equation*}
% \label{eq:inference}
    \bm a_i = \argmax_{x_i} \sum_{x_{i} \in \mathcal{V}(\mathcal{G}_{q_i})} \exp\left(\sum_{q_j \in \mathrm{kNN}_{q_i}} \mathrm{sim}(\bm x_i, \mathcal{A}_j)\right)
\end{equation*}

\section{Experiments}
\label{sec:experiments}
In this section, we demonstrate the effectiveness of the semiparametric approach of \alg and show that the nonparametric and parametric component offer complementary strengths. 
For example, we show that the model performance improves as more evidence is dynamically retrieved by the nonparametric component (\S\ref{sub:knn_increase}).
Similarly, \alg can handle queries requiring reasoning patterns more complex than simple chains (i.e. subgraphs) because of the inductive bias provided by GNNs (\S\ref{sub:subgraph_pattern}). 
It can handle new and unseen entities because of the sparse entity input features as a part of its design (\S\ref{sub:subgraph_pattern}). 
We also show that the nonparametric subgraph selection of \alg allows us to operate over a massive real-world KG (full Freebase KG) and obtain very competitive performance on several benchmark datasets including WebQuestionsSP \cite{yih2016value}, FreebaseQA \cite{jiang2019freebaseqa} and MetaQA \cite{zhang2018variational}.

\subsection{Reasoning over Complex Patterns}
\label{sub:subgraph_pattern}
We want to test whether \alg can answer queries requiring complex reasoning patterns.
Note that, the reasoning patterns are always latent to the model, i.e. the model has to answer a given query from the query-subgraph and the retrieved KNN-subgraphs \emph{without} any knowledge of the structure of the pattern.

\begin{figure}
    \centering
    \includegraphics[width=\columnwidth]{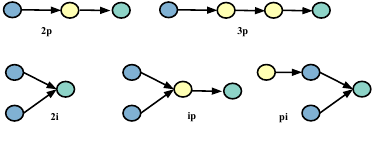}
    \vspace{-10mm}
    \caption{Various reasoning pattern types considered.}
    \label{fig:shapes_patterns}
    \vspace{-3mm}
\end{figure}

To test the model capacity to identify reasoning patterns, we devise a controlled setting in which the model has to infer reasoning patterns of various shapes (Figure~\ref{fig:shapes_patterns}), inspired by \citet{ren2020query2box}. 
Note that in their work, the task was to execute the input structured query on an incomplete KB, i.e, the shape of the input patterns are known to the model. 
In contrast, in our setting, the model has to find the answer node (marked \tikzcircle[fill=mygreen]{3pt}), which is nestled in each of the structured pattern without the knowledge of the pattern structure. 
Also note, there can be multiple nodes of same type as the answer type, so the task cannot be completed by solving easier task of determining entity types. 
Instead the model has to identify specific \tikzcircle[fill=mygreen]{3pt} nodes which are at the end of reasoning patterns (there can be multiple \tikzcircle[fill=mygreen]{3pt} nodes in the graph).

\textbf{Data Generation Process.} 
We first define a type system with a set of entity types $\mathcal{E}$ and relation types $\mathcal{R}$. 
The type-system also specifies a set of `allowed relation types' between different pairs of entity types. 
For example, an `employee' KB relation is defined between an `organization' and `people' entity types. 
Entities (or nodes $\mathcal{V}$) are then generated uniformly from the set of entity types $\mathcal{E}$. 
Next, relation edges (uniformly sampled from the allowed types) are joined between a pair of nodes with a probability $p$ following the Erd\H{o}s-R\'enyi model \cite{erdos1960evolution} of random graph generation. 
To ensure that models only rely on the graph structure, each graph has a `unique' set of entities and no two graphs share entities. 
This also effectively tests how much the nonparametric property of \alg can reason with unseen entities.
More details regarding the hyperparameters $\mathcal{E, R, V}$ are included in the appendix \ref{sub:synthetic_sampling}.

\begin{figure}
    \centering
    \includegraphics[width=\columnwidth]{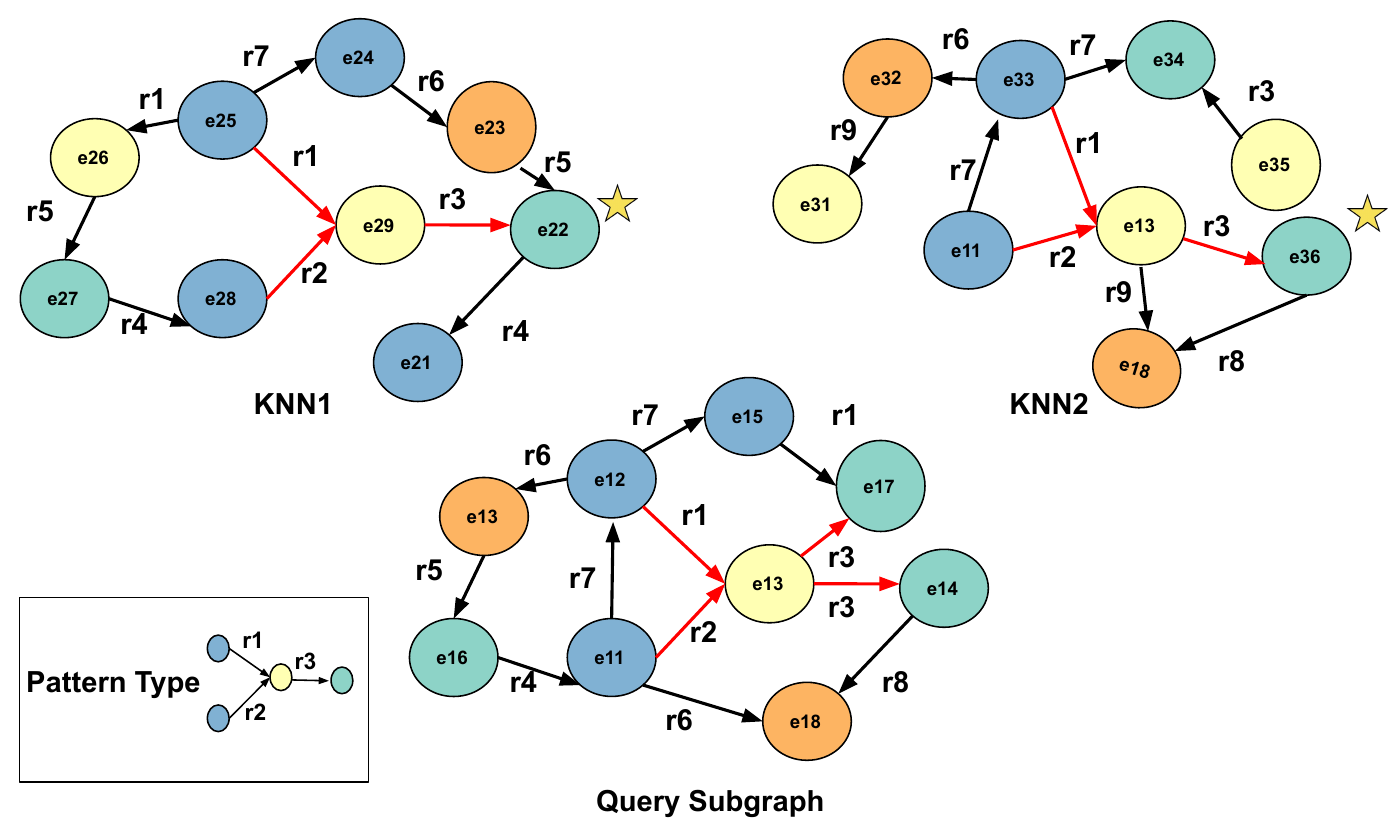}
    \vspace{-5mm}
    \caption{Setup for reasoning over subgraph patterns. Note that the same `pattern type' repeats across subgraphs. Also note, that the query graph (bottom) has two answer nodes (e14, e17). Since every subgraph has its own set of unique entities, hence a model has to reason over the similarities in graph structures to find the answer node in the query-subgraph.}
    \label{fig:synthetic_task}
\end{figure}

\textbf{Pattern Generation}. A pattern is next sampled from the set of shapes shown in Figure~\ref{fig:shapes_patterns}. 
The sampled pattern merely suggests the structure of the desired pattern. `\emph{Grounding}' a pattern shape involves assigning each nodes with an entity present in a generated graph. 
Similarly each edge of the pattern type is assigned a relation from the set of allowed relation types. 
After grounding the pattern, we ``insert'' the pattern in the graph. Since the nodes of the grounded pattern already exists in the graph, inserting a pattern in the graph amounts to adding the edges of the grounded pattern to the graph that already did not exist in it. 
We also define a `pattern type' --- that refers to a pattern whose edges have been assigned relation types but the nodes have not been assigned to specific entities (bottom-left corner in Figure~\ref{fig:synthetic_task}). 
Each pattern type is assigned an identifier and queries with the same pattern type are grouped together.

We generate 1000 graphs in each of the training, validation and test sets. 
We generate 200 pattern types whose shapes are uniformly sampled from the 5 shapes shown in Figure~\ref{fig:shapes_patterns}. 
Therefore, there are around 40 examples of each pattern shape and 
around 5 examples of each pattern type. 
This is consistent with real-world setting where a model will encounter a reasoning pattern only very few times during training. 
For a query with a particular pattern type, other training queries with the same pattern type form its nearest neighbors.

\textbf{Baselines.} Because of the inductive nature of this task where only new entities are seen at test time, most parametric KG reasoning algorithms \cite{bordes2013translating,yang2014embedding,sun2019rotate} will not work out of the box. 
We extend the widely used KG reasoning model --- TransE \cite{bordes2013translating} to work in the inductive setting. 
Specifically, instead of having a fixed vocabulary of entities, the objective function is computed on the dense representation obtained from the output of the GNN layers. 
This also makes the comparision with \alg fair since it also operates on the same representations albeit with a contrastive loss. 
KG completion algorithms also need a query-relation as input. Each pattern type for a query serves as the query relation. 
Apart from the parametric baseline, we also test a simple nonparametric approach, CBR-path in which path patterns which connect source and target entities are gathered and applied for the current query. 
Comparing \alg to CBR-path will help us understand the importance of modeling subgraph patterns rather than simple chains. 

\textbf{Does \alg have the right inductive bias?}
The first research question that we try to answer is, if \alg has the right inductive bias for this task. 
We test \alg that has undergone \emph{no training}, i.e. the parameters of the GNN are randomly initialized. 
Note the model still takes as input the sparse representation of entities. 
This experiment will help us answer if the node representations actually capture the local structure around them and whether the answer node can be found by doing a search w.r.t the answer nodes in the KNN-query subgraphs.

\begin{table}[t]
    \centering
    \scalebox{0.84}{
    \begin{tabular}{@{}l c c c c c c@{}}
    \toprule
    Model & \bf 2p & \bf 3p & \bf 2i & \bf ip& \bf pi & \bf avg.  \\
    \midrule
    \alg (NT) & 68.56 & 84.35& 23.00 & 34.85 & 35.35 & 47.28\\
    GNN + TransE & 80.03 & 74.49 & 80.00 & 52.67 & 81.53 & 72.69\\
    CBR-path & 69.71 & 54.39 & \textbf{100.00} & 69.12 & 51.24 & 71.09 \\
    \alg & \textbf{96.64} & \textbf{88.43} & 90.46 & \textbf{70.02} & \textbf{86.81} & \textbf{85.68}\\
    \bottomrule
    \end{tabular}}
    \caption{Strict Hits\@1(\%) for predicting \emph{all} the answer nodes correctly. \alg (NT) denotes using \alg with no training and we see that it performs decently suggesting that it has the right inductive bias for the task. On training, the performance improves on all subgraph patterns.}
    \label{tab:results_subgraph_patterns}
\end{table}

Table~\ref{tab:results_subgraph_patterns} reports the strict hits@1 on this task, i.e. to score a query correctly, a model has to identify and rank \emph{all} answer nodes above all other nodes in the graph. 
The first row of Table~\ref{tab:results_subgraph_patterns} shows the results. For comparison, a random performance on this task is $\frac{1}{\lvert\mathcal{V}\rvert} = \frac{1}{120} = 0.83\%$. 
As it is clear from the results, an un-trained \alg achieves performance much higher than random performance. 
Its quite high for the simple 2p and 3p patterns. For other patterns that need the more complicated intersection operation, the performance degrades, but is still much higher than random.

\textbf{Our Results.} On training \alg, the performance of the model drastically improves for each pattern type reaching an average performance of 85.68\%. 
The performance on pattern types which are more complex than chains (ip, pi) etc are worse than chain-type patterns (2p, 3p) suggesting that our task is non-trivial. \\
\textbf{On comparison to parametric model.} 
This experiment helps us understand whether a model can learn to memorize and store patterns effectively (for each query relations) when it has seen few examples of that pattern during training. 
Row 2 of Table~\ref{tab:results_subgraph_patterns} shows the performance of GNN + TransE model. 
We find that the parametric model performs worse than \alg on all the query types reaching an average performance of 13\% point below \alg. 
This shows that a semiparametric model with a nonparametric component that retrieves similar queries at inference can make it easier for the model to reason effectively.
In practice, we had to train this model for a much longer time than training \alg.\\
\textbf{On comparison to path-based model. } 
From Table~\ref{tab:results_subgraph_patterns}, we can see that \alg outperforms CBR-path by more than 14\% points suggesting that reasoning over subgraphs is a more powerful approach that reasoning with each paths independently. 
On the `2i' pattern, CBR-path outperforms \alg  since `2i' can be seen as 2 independent paths intersecting at one node and CBR is able to model that perfectly.
However, when the pattern needs composition and intersection and path-traversal, CBR-path struggles and performs much worse.

\begin{table}[t]
\centering
\scalebox{0.95}{
\begin{tabular}{@{}l c c c c@{}}
\toprule
\multirow{2}{*}{\textbf{Model}} & \multicolumn{3}{c}{\textbf{MetaQA}} & \multirow{2}{*}{\textbf{WebQSP}} \\ 
\cmidrule{2-4}
   & \textbf{1-hop}   &   \textbf{2-hop}   & \textbf{3-hop}   &    \\
\midrule
KVMemNN{\scriptsize~\cite{miller2016key}}  &  95.8  &  25.1  &  10.1  &   46.7 \\
GraftNet{\scriptsize~\cite{sun2018open}}  &  97.0  &  94.8  &  77.7  &  66.4   \\
PullNet{\scriptsize~\cite{sun2019pullnet}}  &  97.0  &  99.9  &  91.4  &  68.1 \\
SRN{\scriptsize~\cite{qiu2020stepwise}}   &   97.0  &   95.1  &   75.2  &  -   \\
ReifKB{\scriptsize~\cite{cohen2020scalable}}  &  96.2  &  81.1  &  72.3  &  52.7  \\
EmbedKGQA{\scriptsize~\cite{saxena2020improving}}  &  \textbf{97.5}  &  98.8  &  94.8  &  66.6   \\
NSM{\scriptsize~\cite{he2021improving}} & 97.2 & 99.9 & 98.9 & \textbf{74.3}\\
\alg (Ours) & 97.1 & 99.8 & \textbf{99.3} & 72.1\\\bottomrule
\end{tabular}}
\vspace{-2mm}
\caption{Performance on \webqsp and \metaqa benchmarks.}
\label{tab:webq_metaqa_results}
\end{table}

\subsection{Performance on benchmark datasets}
\label{sub:benchmark_performance}
Next, we test the performance of \alg on various KBQA benchmarks --- \metaqa \cite{zhang2018variational}, \webqsp \cite{yih2016value} and \fbqa \cite{jiang2019freebaseqa}. \metaqa comes with its own KB. 
For other datasets, the underlying KB is the full Freebase KB containing over 45 million entities (nodes) and 3 billion facts (edges). 
Please refer to the appendix for details about each dataset (\S\ref{sub:datasetdetails}).

Our main baselines are the two semiparametric models that provide both a mechanism to gather query subgraphs for a given query and reason over them to find the answer --- GraftNet \cite{sun2018open}, PullNet \cite{sun2019pullnet}. 
GraftNet uses personalized page rank to determine which edges are relevant for a particular query and PullNet uses a multi-step retriever that at each step, classifies if an edge is relevant to the current representation of the query. 
For their reasoning model, both works use a graph convolution model and treat the answer prediction as a node classification task. 
However unlike us, they do not use query-subgraphs of KNN queries. 
Followup KBQA works \citep[inter-alia]{saxena2020improving,he2021improving} use the query-specific graphs provided by GraftNet from their open-source code and do not provide a mechanism to gather query-specific subgraphs. 
However, for completeness, we report and compare with those methods as well.

Table~\ref{tab:webq_metaqa_results} reports the performance on \webqsp and all three partitions of \metaqa. 
When compared to GraftNet and PullNet, \alg performs much better on an average on both the datasets. 
On the more challenging 3-hop subset of MetaQA, \alg outperforms PullNet by more than 7 points and GraftNet by more than 15 points. 
This shows that even though these two models use a GNN for reasoning, using information from subgraphs from similar KNN queries leads to much better performance.
On \webqsp, we outperform all models except the recently proposed NSM model \cite{he2021improving}. 
But as we noted before, NSM operates on the subgraph created by GraftNet and does not provide any particular mechanism to create its own query-specific subgraph (an important contribution of our model). 
Moreover NSM is a parametric model and will not have some advantages of nonparametric architectures such as ability to handle new entities and reasoning with more data.
Table~\ref{tab:freebase_qa_results} reports the results on the \fbqa dataset, which contains real trivia questions obtained from various trivia competitions. 
Thus the questions can be challenging in nature. We compare with other KBQA models reported in \citet{han-etal-2020-empirical}. 
Most of the models are pipelined KBQA systems that rely on relation extraction to map the query into a KB edge. 
\alg outperforms all the models by a large margin. 
We also report the performance on two models that use large LMs and large-scale pre-training. 
\alg, which only operates on the KB has a performance very close to the performance of Entity-as-Experts model \cite{fevry2020entities}. 
We leave the integration of large LMs into our parametric reasoning component as future work.

\begin{table}[t]
    \centering
    \hspace{9mm}
    \begin{tabular}{@{}l c@{}}
    \toprule
    Model & Accuracy \\
    \midrule
    \textit{KB-only models} & \\\midrule
    HR-BiLSTM \cite{yu2017improved} & 28.40 \\
    KBQA-Adapter \cite{wu2019learning} & 28.78\\
    KEQA \cite{huang2019knowledge} & 28.73\\
    FOFE \cite{jiang2019freebaseqa} & 37.00\\
    BuboQA \cite{mohammed2017strong} & 38.25\\
    \alg (Ours) & \textbf{52.07}\\
    \midrule
    \textit{LM pre-training + KB} & \\
    \midrule
    EAE \cite{fevry2020entities} & 53.4 \\
    FAE \cite{verga2020facts} & \textbf{63.3} \\
    \bottomrule
    \end{tabular}
    \caption{Top-1 \% accuracy on the \fbqa dataset. The top section reports performance of models that operate only on KBs. The bottom section reports performance on models that also use additional knowledge stored in large language models.}
    \label{tab:freebase_qa_results}
\end{table}

\begin{table}[b]
    \centering
    \small
    \begin{tabular}{@{}l c c c c@{}}\toprule
         Subgraph & \#edges &  \#relations & \#entities & coverage(\%) \\\midrule
         \textit{\webqsp} & & & &  \\\midrule
         Graft-net & 4306.00 & 294.69 & 1447.68 & 89.93\%\\
         \alg & 1934.65 & 36.42 & 1403.87 & 94.30\%\\
         \% diff & -55.07\% & -87.64\% & -3.02\%& +4.85\%\\\midrule
         \textit{\metaqa-2} & & & &  \\\midrule
         Graft-net & 1126.0 & 18.00 & 468.00 & 99\%\\
         \alg & 89.21 & 4.72 & 77.52 & 99.9\%\\
         \% diff & -92.07\% & -73.78\% & -83.43\% & +0.91\% \\\bottomrule
    \end{tabular}
    \caption{Our adaptive subgraph collection strategy produces a compact subgraph for a query while increasing recall.}
    \label{tab:graph_stats}
\end{table}

\subsection{Analysis}
\label{sub:knn_increase}

\textbf{How effective is our adaptive subgraph collection strategy?} Table~\ref{tab:graph_stats} reports few average graph statistics for the query-subgraphs collected by our graph-collection strategy. 
We also compare to GraftNet's subgraphs. 
As can be seen, our adaptive graph collection strategy produces much more compact and smaller graphs \emph{while} increasing recall of answers. 
We also consistently find that our graph contains relations which is more relevant to the questions than the subgraph produced by GraftNet (\S\ref{sub:adaptive_v_graft}).
Table~\ref{tab:adaptive_subgraph_comparison} reports the performance of \alg when trained and tested on the subgraph obtained from Graftnet and our adaptive procedure, demonstrating the effectiveness of our adaptive subgraph collection method.

\begin{figure}[t]
    \centering
    \includegraphics[width=0.6\columnwidth]{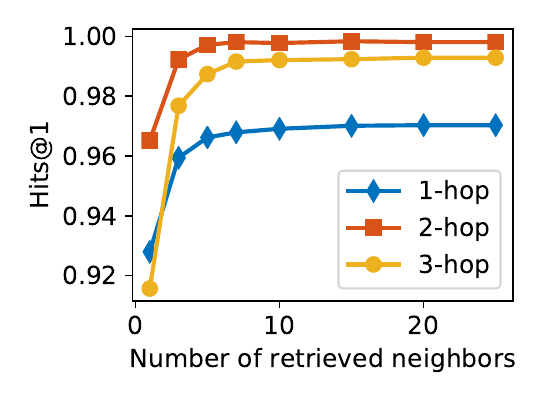}
    \vspace{-3mm}
    \caption{Performance of \alg as more nearest neighbors are introduced at test-time. Results on different partitions of \metaqa.}
    \label{fig:increase_knn}
\end{figure}

\begin{table}[ht]
    \centering
    \small
    \begin{tabular}{ l  c  c }\toprule
        Subgraph & WebQSP & MetaQA-3 \\\midrule
        GraftNet & 65.61\% & 96.90\% \\
        Adaptive & 72.10\% & 99.30\%\\\bottomrule
    \end{tabular}
    \caption{Performance of \alg with adaptive subgraph and GraftNet subgraph}
    \label{tab:adaptive_subgraph_comparison}
\end{table}

\begin{table}
    \centering
    \scalebox{0.9}{
    \begin{tabular}{c c c c c c c c}\toprule
     & 1 & 2 & 3 & 5 & 7 & 10 & 20\\\midrule
    Acc & 69.06 & 70.28 & 71.20 & 72.11 & 71.14 & 70.71 & 69.12\\\bottomrule
    \end{tabular}}
    \caption{Performance on WebQSP w.r.t varying number of KNN questions. Unlike MetaQA, the performance decreases on adding KNNs beyond a certain point as irrelevant questions are retrieved because of the limited size of WebQSP training set. }
    \label{tab:knn_webqsp}
\end{table}

\textbf{Can \alg reason with more evidence?}
A desirable property of nonparametric models is to be able to `improve' its prediction as more evidence is made available. 
We test \alg by taking a trained model and issuing it an increasing number of nearest neighbor queries. As we see from Figure~\ref{fig:increase_knn}, the performance of \alg drastically improves as we increase the number of nearest neighbors from 1 to 7 and then increases at a lower rate and converges at around 10 nearest neighbors. 
This is because, the model has all the required information it needs from its nearest neighbors.
However, in the much smaller WebQSP dataset we observe a different behavior (Table~\ref{tab:knn_webqsp}). 
This is because of the limited size of the dataset, irrelevant questions starts appearing in the context as we increase the number of KNNs beyond a certain limit.

\begin{figure}[t]
    \centering
    \includegraphics[width=0.86\columnwidth]{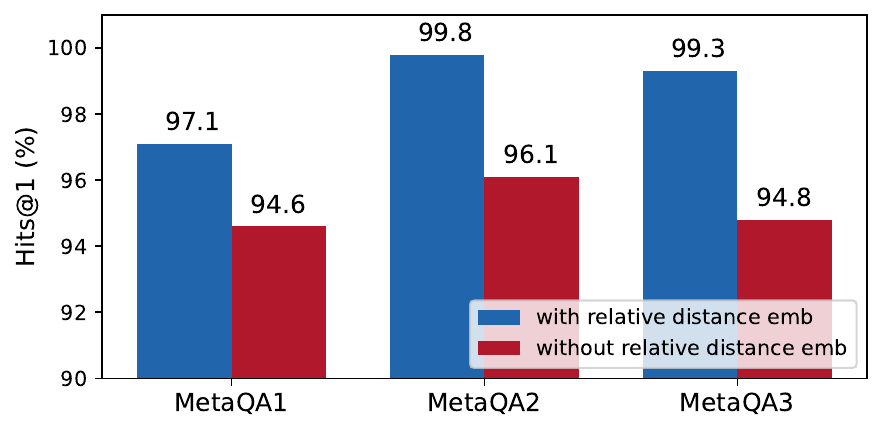}
    \vspace{-3mm}
    \caption{Performance with and without relative distance embedding. The relative distance from the query entity clearly is important for achieving good performance.}
    \label{fig:distance_emb_ablation}
\end{figure}

\noindent\textbf{Are relative distance embeddings important?} Figure~\ref{fig:distance_emb_ablation} shows the performance of \alg with and without the relative distance embeddings (\S\ref{sub:gnn}). 
It is clear that capturing the relative distance from the query entities provide serves as a helpful feature for the model. 

\section{Conclusion}
\label{sec:conc}
In this work, we explored a semiparametric approach for KBQA.
We demonstrated \alg poses several desirable properties approach in which nonparametric and parametric component offer complementary strengths.
By retrieving similar queries and utilizing the similarities in graph structure of local subgraphs to answer a query, our approach is able to handle complex questions as well as generalize to new types of questions.
Exploring different types of parametric models with different reasoning capabilities (LMs, GNNs, etc.) would be an interesting future research direction.
Another avenue of potential research would be a never-ending learning type of system where we keeps adding newly discovered facts in the nonparametric part.
\section*{Acknowledgments}
This work is funded in part by the IBM Research AI through the AI Horizons Network, the Chan Zuckerberg Initiative under the project Scientific Knowledge Base Construction, the National Science Foundation under Grant Number IIS-1763618, the National Science Foundation under Grant Number IIS-1922090, the Defense Advanced Research Projects Agency via Contract No. FA8750-17-C-0106 under Subaward No. 89341790 from the University of Southern California, and the Office of Naval Research via Contract No. N660011924032 under Subaward No. 123875727 from the University of Southern California. The work reported here was performed using high performance computing equipment obtained under a grant from the Collaborative R\&D Fund managed by the Massachusetts Technology Collaborative.

\bibliography{example_paper}
\bibliographystyle{icml2022}

%%%%%%%%%%%%%%%%%%%%%%%%%%%%%%%%%%%%%%%%%%%%%%%%%%%%%%%%%%%%%%%%%%%%%%%%%%%%%%%
%%%%%%%%%%%%%%%%%%%%%%%%%%%%%%%%%%%%%%%%%%%%%%%%%%%%%%%%%%%%%%%%%%%%%%%%%%%%%%%
% APPENDIX
%%%%%%%%%%%%%%%%%%%%%%%%%%%%%%%%%%%%%%%%%%%%%%%%%%%%%%%%%%%%%%%%%%%%%%%%%%%%%%%
%%%%%%%%%%%%%%%%%%%%%%%%%%%%%%%%%%%%%%%%%%%%%%%%%%%%%%%%%%%%%%%%%%%%%%%%%%%%%%%
\newpage
\appendix
\onecolumn
\section{Hyperparameters}
For MetaQA, we use 3 GCN layers with GCN layer dimension of 32. For training we have used 5 nearest neighbors and 10 are used for evaluation for the 1-hop, 2-hop and 3-hop queries.We optimize the loss using Adam Optimizer with beta1 of 0.9, beta2 of 0.999 and epsilon of 1e-8. As well as the learning rate is set to be 0.00099 with temperature value of 0.0382 (1-hop), 0.0628 (2-hop) ,0.0779 (3-hop). All the models are trained for 5 epochs.

Similarly for WebQSP, we use 3 GCN layers with GCN layer dimension of 32. But for training we used 10 nearest neighbors and 5 are used for evaluation. We optimize the loss using Adam Optimizer with beta1 of 0.9, beta2 of 0.999 and epsilon of 1e-8. As well as a learning rate of 0.0024 and temperature of 0.0645 is used. The model is trained for about 30 epochs. All hyper-parameters can also be found in our code-base.
%%%%%%%%%%%%%%%%%%%%%%%%%%%%%%%%%%%%%%%%%%%%%%%%%%%%%%%%%%%%%%%%%%%%%%%%%%%%%%%

\section{Generating synthetic data for control experiments}
\label{sub:synthetic_sampling}

We generate the dataset for running control experiments extending the Erd\H{o}s-R\'enyi model \cite{erdos1960evolution} for sampling random graphs to heterogeneous graphs (graphs with types edges and/or nodes).

(i) In the first stage, a type system for the KB is created by sampling a fixed set of $16$ entity types and edges are added between types with a chance of $0.3$. Our sampled KB type system has $74$ relation types. This is the exact Erd\H{o}s-R\'enyi model. 

(ii) Next, given a pattern shape we generate a `grounded pattern'. The first query entity is selected at random. From there every entity type/relation in the pattern type is sampled from the types allowed by the KB system. For example, given a $2i$ pattern shape, we sample an entity type $t1$ for the first query entity $e1$. Then we assign a type $r1$ to outgoing edge from the allowed outgoing edge types for $t0$. This assigns a type $t_ans$ to the answer node $?ans$. Then we sample a type $r2$ to incoming edge from the allowed incoming edge types for $t_ans$. This assigns a type $t2$ to the second query entity $e2$. The final `grounded' $2i$ pattern is then (($e1$, $r1$, $?ans$), ($e2$, $r2$, $?ans$)).

(iii) Next, to sample a query graph, we create an empty graph with $120$ entities each randomly assigned a type from the $16$ types. The query entities have pre-assigned entity types based on the pattern type ($e1$ and $e2$ from the previous example are fixed to be type $t1$ and $t2$ respectively) . Starting from the query entities, we sample edges allowed by the KB type system. We add an edge between two entities with a chance of $0.4$. We ensure that the entities in the subgraph are at most a distance of 3-hops from the query entities.

(iv) Finally, we execute the pattern on the graph to assign labels to answer nodes.

Our control dataset samples 200 pattern types and 15 graphs per pattern type distributing them equally (i.e. 5 each) between train, validation and test. For each of the 15 graphs that share a common pattern type, we assign the 5 graphs that were put in the train set as the kNN queries.

\section{Dataset details and statistics}
\label{sub:datasetdetails}
Table~\ref{tab:dataset_stats} summarizes the basic statistics of the datasets used in our experiments.
\begin{table}[h]
    \centering
    \small
    \begin{tabular}{l c c c c}\toprule
         Dataset & Train & Dev & Test \\\midrule
         MetaQA 1-hop & 96,106 & 9,992 & 9,947\\
         MetaQA 2-hop & 118,980 & 14,872 & 14,872 \\
         MetaQA 3-hop & 114,196 & 14,274 & 14,274\\
         WebQSP & 2,848 & 250 & 1,639 \\
         FreebaseQA & 20,358 & 2308 & 3996\\\bottomrule         
    \end{tabular}
    \caption{Dataset Statistics}
    \label{tab:dataset_stats}
\end{table}

%%%%%%%%%%%%%%%%%%%%%%%%%%%%%%%%%%%%%%%%%%%%%%%%%%%%%%%%%%%%%%%%%%%%%%%%%%%%%%%
\section{Retrieving cases by masking query entities}

\begin{table}[H]
    \centering
    \scriptsize
    \begin{tabular}{l | l | l}\toprule
        & KNN for Unmasked Query & KNN for Masked Query \\\midrule
        Query & what did \textbf{james k polk} do before he was president & what did [MASK] do before he was president \\\hline
        \multirow{2}{*}{Retrieved kNN} & 1. what did \textbf{james k polk} believe in & 1. what did abraham lincoln do before he was president \\
        & 2. what did barack obama do before he took office & 2. what did barack obama do before he took office \\\midrule
        Query & what are the songs that \textbf{justin bieber} wrote & what are the songs that [MASK] wrote \\\hline
        \multirow{3}{*}{Retrieved kNN} & 1. what is the name of \textbf{justin bieber} brother & 1. what are all the songs nicki minaj is in \\
        & 2. what are all the inventions benjamin franklin made & 2. what songs did mozart write \\
        & 3. what are all the movies channing tatum has been in & 3. what songs did richard marx write \\\midrule
        Query & where did edgar allan poe died & where did [MASK] died \\\hline
        \multirow{3}{*}{Retrieved kNN} & 1. what college did \textbf{edgar allan poe} go to & 1. where did mendeleev died \\
        & 2. what magazine did \textbf{edgar allan poe} work for & 2. where did benjamin franklin died \\
        & 3. what year did \textbf{edgar allan poe} go to college & 3. where did thomas jefferson died
        \\\bottomrule
    \end{tabular}
    \caption{Retrieval by masking the question entity prevents the returned kNN queries from focusing on the entity and instead rely on the question structure and relation involved.}
    \label{tab:eye_candy}
\end{table}

Table~\ref{tab:eye_candy} shows example of query retrieval when the entity mentions in the input query is masked or not. Since \alg prefers KNN queries that have more relational similarity and not necessarily about the same entity in the question, it is clear that masking of query helps in retrieving more relevant questions.

\section{Adaptive subgraph collection tailors subgraphs to the query}
\label{sub:adaptive_v_graft}
\begin{figure}[t]
    \centering
    \small
    \begin{subfigure}[]{0.95\columnwidth}
   \includegraphics[width=0.95\columnwidth]{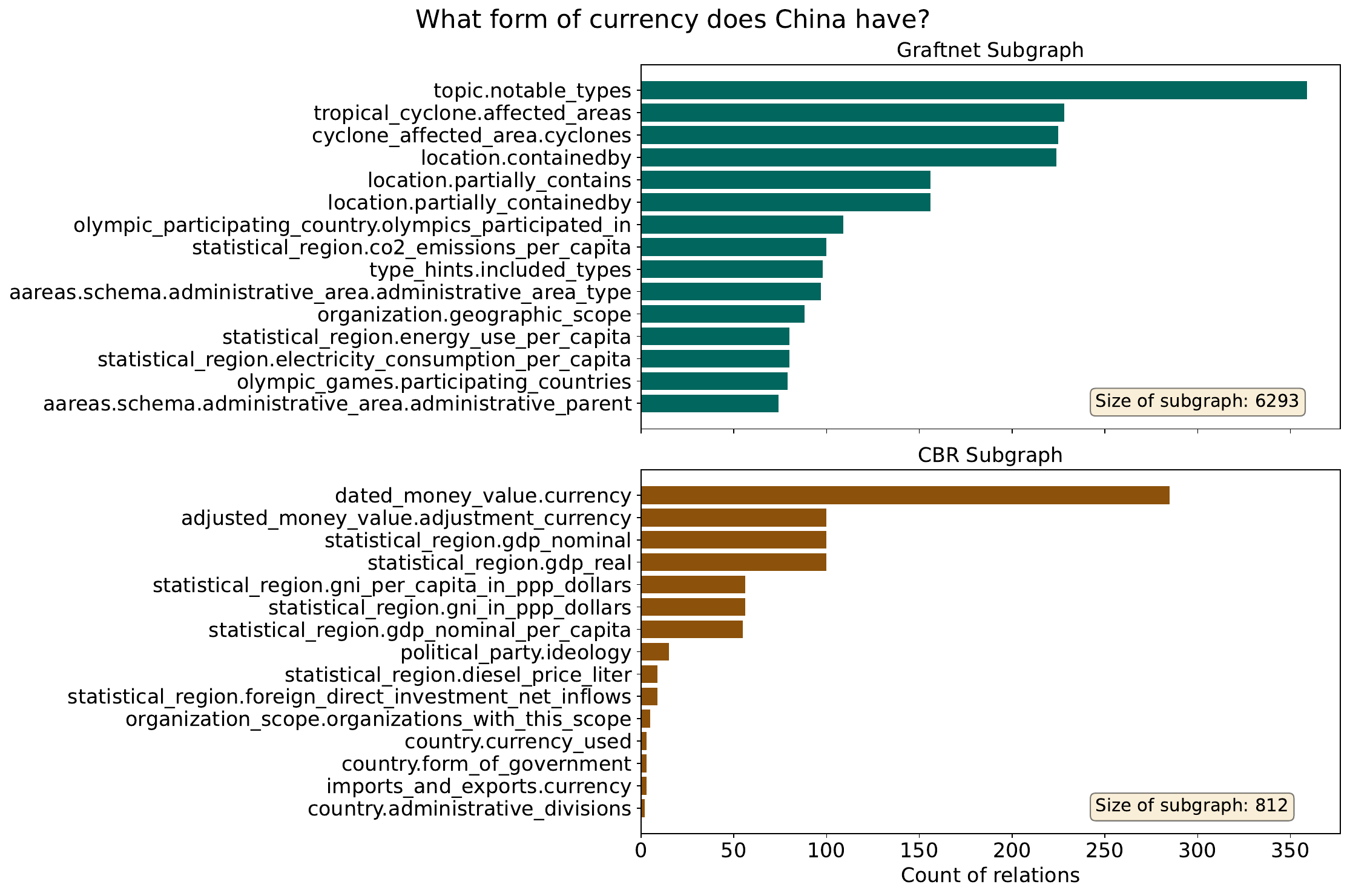}
   \label{fig:Ng1} 
    \end{subfigure}
    \begin{subfigure}[]{0.95\columnwidth}
   \includegraphics[width=0.95\columnwidth]{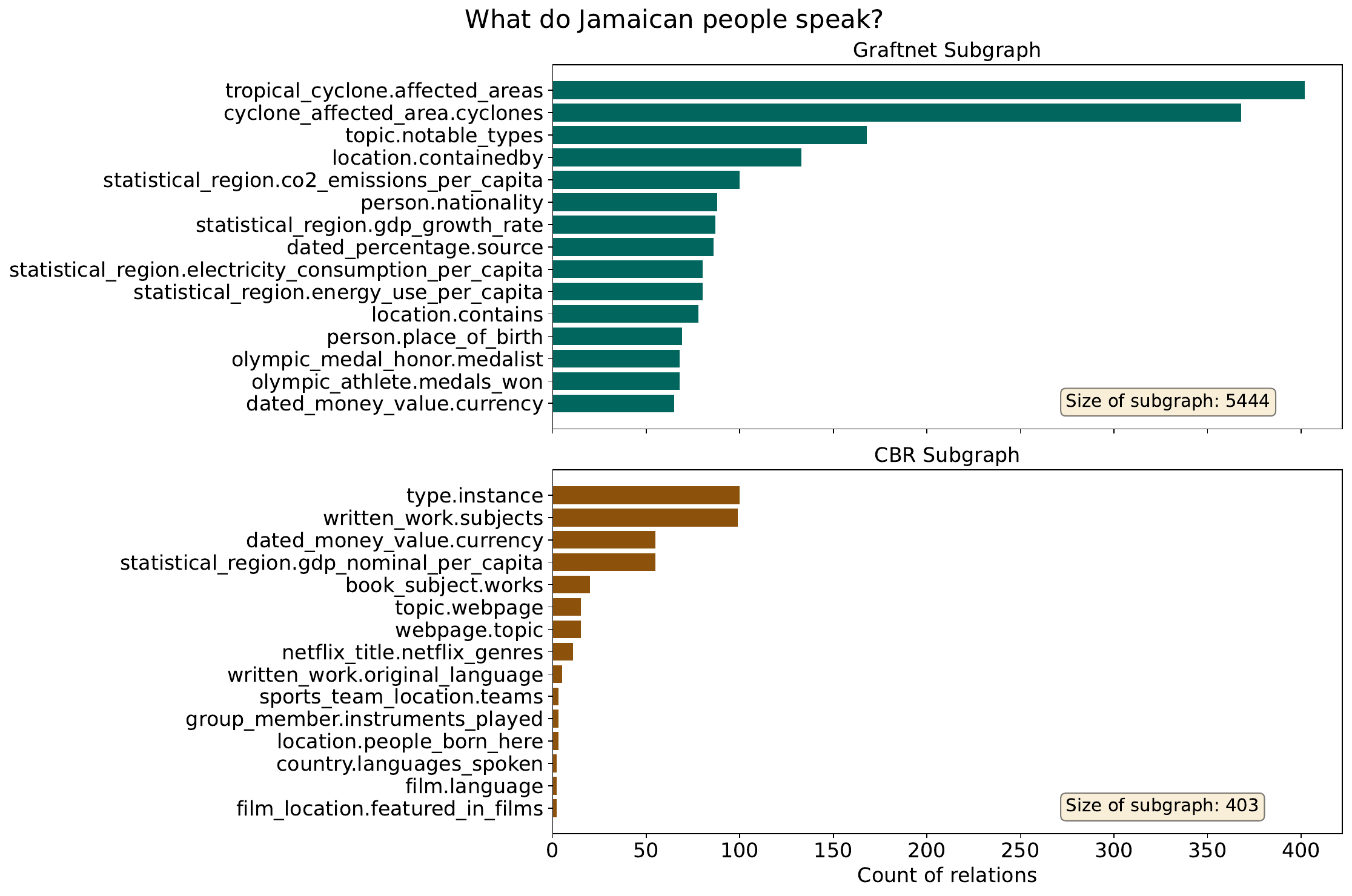}
   \label{fig:Ng2}
    \end{subfigure}
    \caption{}
    \label{fig:adaptive_1}
\end{figure}

\begin{figure}[t]
    \centering
    \small
    \begin{subfigure}[]{0.95\columnwidth}
   \includegraphics[width=0.95\columnwidth]{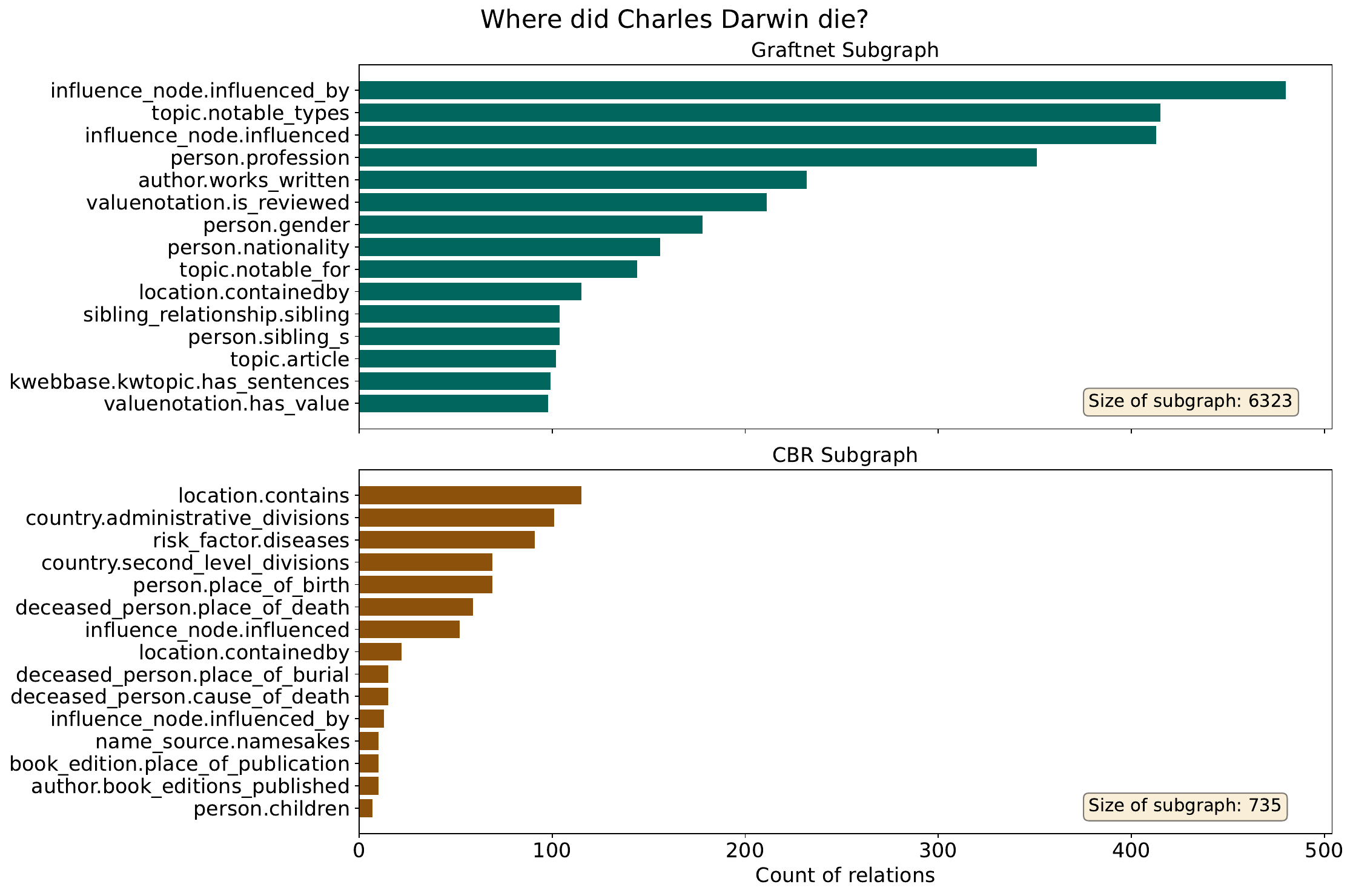}
   \caption{}
   \label{fig:Ng3}
    \end{subfigure}
    
\begin{subfigure}[]{0.95\columnwidth}
   \includegraphics[width=0.95\columnwidth]{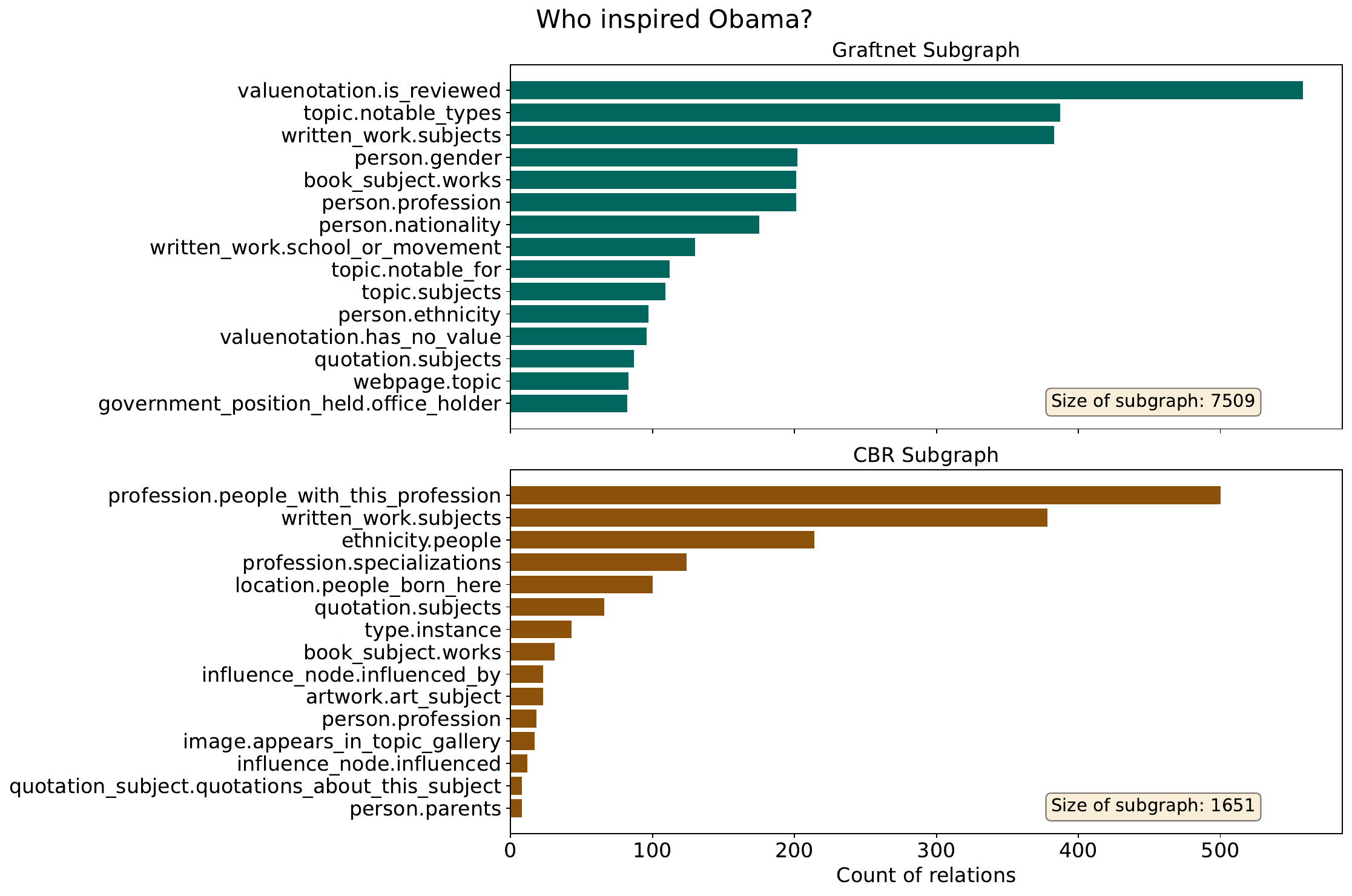}
   \caption{}
   \label{fig:Ng4}
    \end{subfigure}
    \caption{}
    \label{fig:adaptive_2}
\end{figure}

\begin{figure}[t]
    \centering
    \small
    \begin{subfigure}[]{0.95\columnwidth}
   \includegraphics[width=0.95\columnwidth]{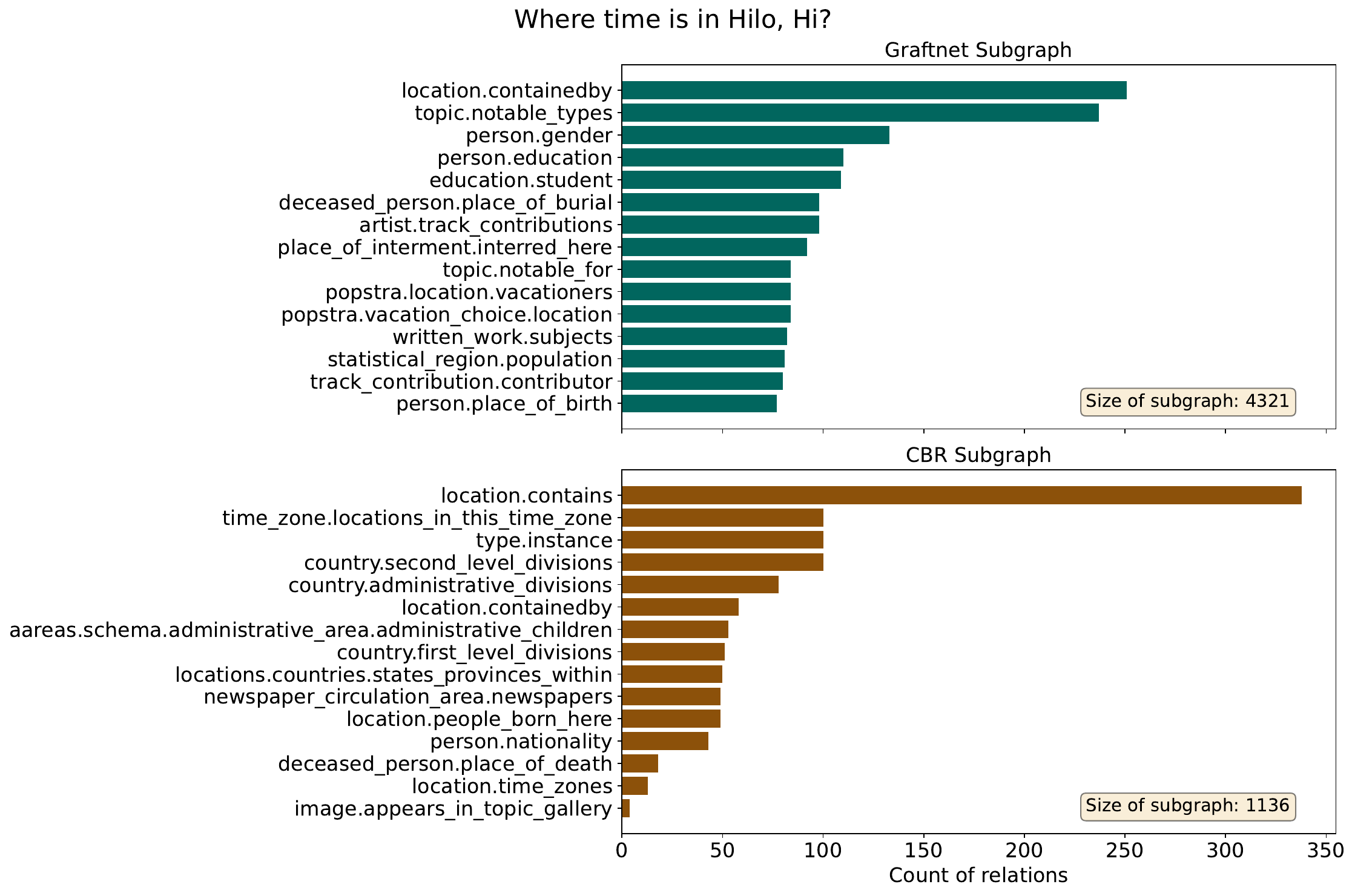}
   \caption{}
   \label{fig:Ng5}
    \end{subfigure}
    
    \begin{subfigure}[]{0.95\columnwidth}
   \includegraphics[width=0.95\columnwidth]{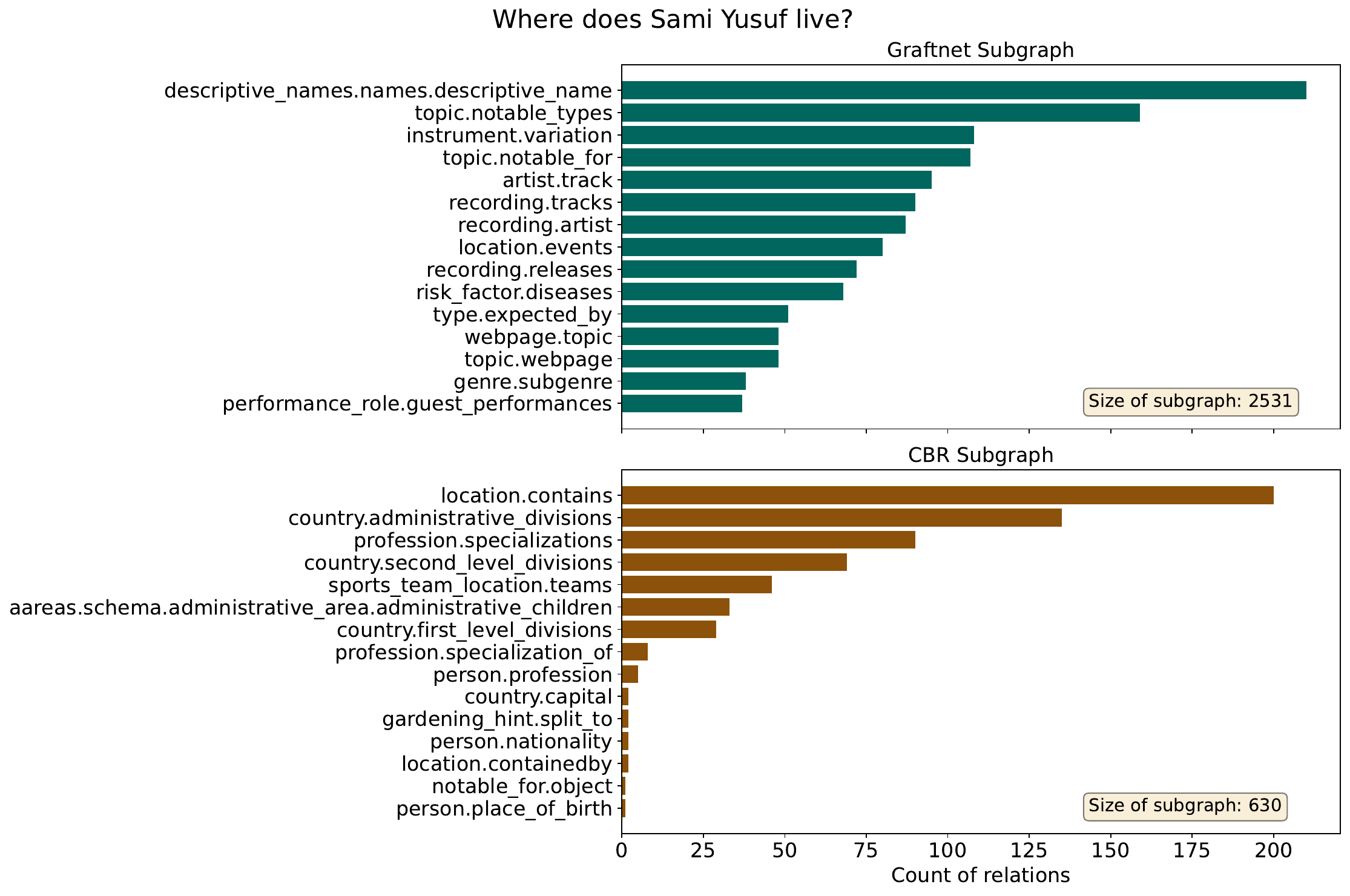}
   \caption{}
   \label{fig:Ng6}
    \end{subfigure}
    \caption{}
    \label{fig:adaptive_3}
\end{figure}

Figure~\ref{fig:adaptive_1}, \ref{fig:adaptive_2} and \ref{fig:adaptive_3} shows example of few query-subgraphs collected by GraftNet and our adaptive subgraph collection strategy (\S\ref{sub:subgraph_retrieval}). Each figure plots the most frequent (top 15) relations gathered by each subgraph collection procedure. 
The size of each subgraph denote the number of edges collected by each subgraph collection procedure.
The subgraph collected by our adaptive strategy produces both compact subgraphs as well as has edges which are more relevant for the query. 
For example, in Figure~\ref{fig:adaptive_1}, for the question "What form of currency does China have?" -- subgraph collected by GraftNet has edges with generic relation types such as "topic.notable\_types", "tropical\_cyclone.affected\_areas" etc, whereas the subgraph collected by our proposed adaptive strategy has edges relevant to answering the question - e.g. "dated\_money\_value.currency". 

% It should be clear that our adaptive subgraph collection strategy gathers relations that are more specific to the query and also leads to more compact subgraph.

\section{Further Related Work}
\label{sec:further_related_work}
\alg shares similarities with the \textsc{retrieve-and-edit} framework \cite{hashimoto2018retrieve} which utilizes retrieved nearest neighbor for structured prediction. 
However, unlike our method they only retrieve a single nearest neighbor and will unlikely be able to generate programs for questions requiring relations from multiple nearest neighbors. 
There has also been a lot of recent work in general NLP which uses KNN-based approaches. 
For example, \citet{khandelwal2019generalization} demonstrate improvements in language modeling by utilizing explicit examples from training data. 
There has been work in machine translation \cite{gu2018search,khandelwal2020nearest} that uses nearest neighbor translation pair to guide the decoding process. 
\end{document}